\documentclass[10pt,twocolumn,letterpaper]{article}

\usepackage{cvpr}
\usepackage{times}
\usepackage{epsfig}
\usepackage{graphicx}
\usepackage{amsmath}
\usepackage{amssymb}
\usepackage{multirow}
\usepackage{booktabs} 

\usepackage[breaklinks=true,bookmarks=false]{hyperref}

\cvprfinalcopy 

\def\cvprPaperID{3655} 

\ifcvprfinal\fi
\begin{document}

\title{Graphonomy: Universal Human Parsing via Graph Transfer Learning}

\author{Ke Gong$^{1,2\dagger}$, \quad Yiming Gao$^{1\dagger}$, \quad Xiaodan Liang$^{1*}$, \\ \quad Xiaohui Shen$^{3}$, \quad Meng Wang$^{4}$, \quad Liang Lin$^{1,2}$\\
\small$^1$Sun Yat-sen University \quad \small$^2$DarkMatter AI Research \small\quad $^3$ByteDance AI Lab \quad \small$^4$Hefei University of Technology \\
   {\tt\small kegong936@gmail.com}, {\tt\small gaoym9@mail2.sysu.edu.cn}, {\tt\small xdliang328@gmail.com}, \\ {\tt\small shenxiaohui@gmail.com}, {\tt\small wangmeng@hfut.edu.cn}, {\tt\small linliang@ieee.org}
}


\newcommand\blfootnote[1]{%
   \begingroup 
   \renewcommand\thefootnote{}\footnote{#1}%
   \addtocounter{footnote}{-1}%
   \endgroup 
}

\maketitle
\begin{abstract}
Prior highly-tuned human parsing models tend to fit towards each dataset in a specific domain or with discrepant label granularity, and can hardly be adapted to other human parsing tasks without extensive re-training. In this paper, we aim to learn a \textbf{single} universal human parsing model that can tackle all kinds of human parsing needs by unifying label annotations from different domains or at various levels of granularity. This poses many fundamental learning challenges, \eg discovering underlying semantic structures among different label granularity, performing proper transfer learning across different image domains, and identifying and utilizing label redundancies across related tasks.

To address these challenges, we propose a new universal human parsing agent, named ``Graphonomy", which incorporates hierarchical graph transfer learning upon the conventional parsing network to encode the underlying label semantic structures and propagate relevant semantic information. In particular, Graphonomy first learns and propagates compact high-level graph representation among the labels within one dataset via Intra-Graph Reasoning, and then transfers semantic information across multiple datasets via Inter-Graph Transfer. Various graph transfer dependencies (\eg, similarity, linguistic knowledge) between different datasets are analyzed and encoded to enhance graph transfer capability. By distilling universal semantic graph representation to each specific task, Graphonomy is able to predict all levels of parsing labels in one system without piling up the complexity. Experimental results show Graphonomy effectively achieves the state-of-the-art results on three human parsing benchmarks as well as advantageous universal human parsing performance.
\end{abstract}

\vspace{-6mm}

\let\thefootnote\relax\footnotetext{$\dagger$ Equal contribution. *Corresponding Author.}

\begin{figure}[t]
\begin{center}
   \includegraphics[width=0.9\linewidth]{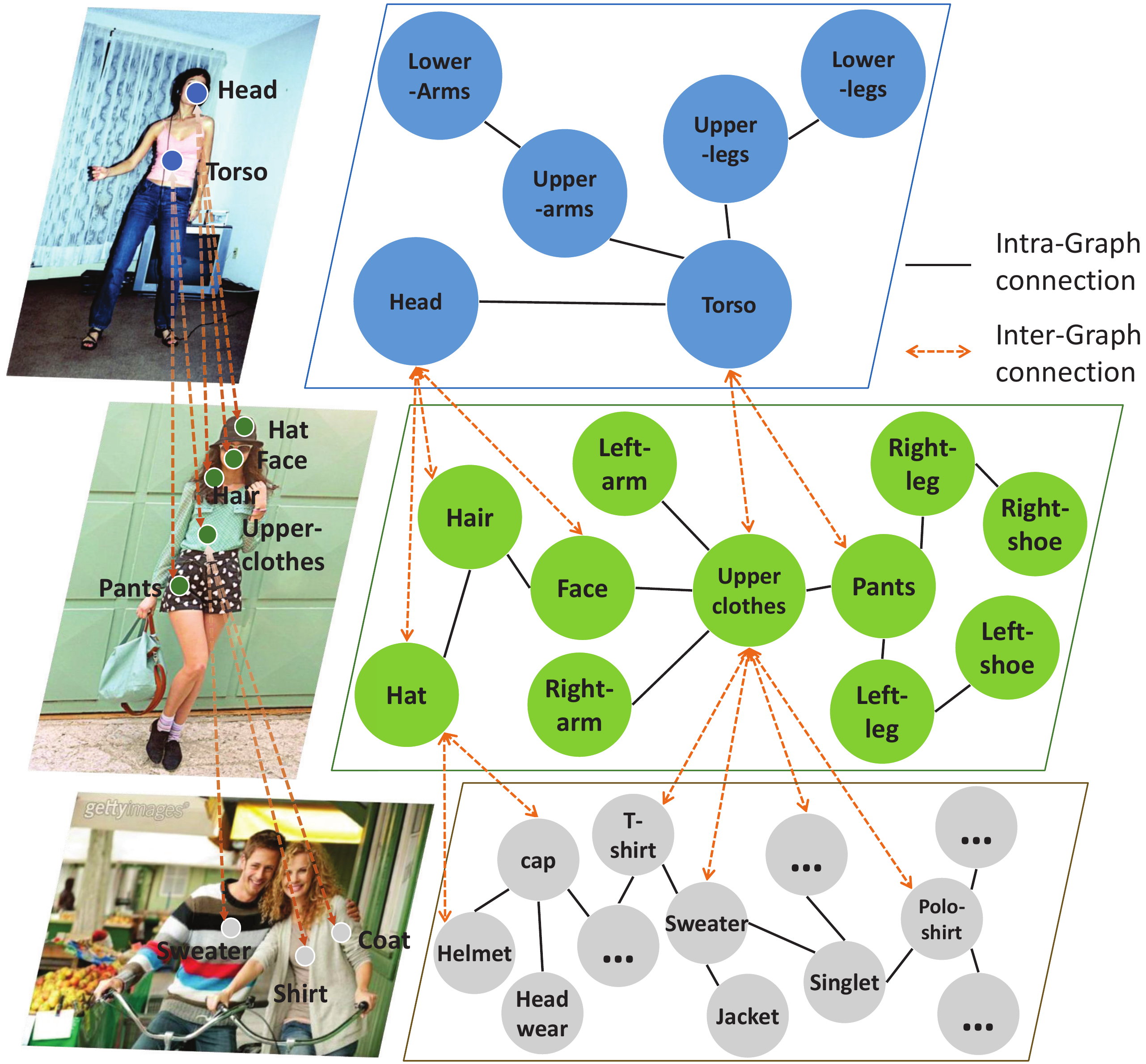}
\end{center}
\vspace{-4mm}
\caption{With huge different granularity and quantity of semantic labels, human parsing is isolated into multiple level tasks that hinder the model generation capability and data annotation utilization. For example, the \textit{head} region on a dataset is further annotated into several fine-grained concepts on another dataset, such as \textit{hat}, \textit{hair} and \textit{face}. However, different semantic parts still have some intrinsic and hierarchical relations (\eg, \textit{Head} includes the \textit{face}. \textit{Face} is next to \textit{hair}), which can be encoding as intra-graph and inter-graph connections for better information propagation. To alleviate the label discrepancy issue and take advantage of their semantic correlations, we introduce a universal human parsing agent, named as ``Graphonomy'', which models the global semantic coherency in multiple domains via graph transfer learning to achieve multiple levels of human parsing tasks.}
\label{fig:graphonomy}
\vspace{-6mm}
\end{figure}

\section{Introduction}
Human visual systems are capable of accomplishing holistic human understanding at a single glance on a person image, e.g., separating the person from the background, understanding the pose, and recognizing the clothes the person wears. Nevertheless, recent research efforts on human understanding have been devoted to developing numerous highly-specific and distinct models for each individual application, e.g. foreground human segmentation task~\cite{Dai_2016_CVPR,He_2017_ICCV}, coarse clothes segmentation task~\cite{ATR,Co-CNN} and fine-grained human part/clothes parsing task~\cite{Gong_2017_CVPR,zhao2018understanding}. Despite the common underlying human structure and shared intrinsic semantic information (e.g. upper-clothes can be interpreted as coat or shirt), these highly-tuned networks have sacrificed the generalization capability by only fitting towards each dataset domain and discrepant label granularity. It is difficult to directly adapt the model trained on one dataset to another related task,  and thus requires redundant heavy data annotation and extensive computation to train each specific model. To address these realistic challenges and avoid training redundant models for correlated tasks, we make the first attempt to investigate a \textit{single} universal human parsing agent that tackles human parsing tasks at different coarse to fine-grained levels, as illustrated in Fig.~\ref{fig:graphonomy}. 
 
The most straightforward solution to universal human parsing would be posing it as a multi-task learning problem, and integrating multiple segmentation branches upon one shared backbone network~\cite{chen2016deeplab,Gong_2017_CVPR,li2017holistic,ATR,Co-CNN}. This line of research only considers the brute-force feature-level information sharing while disregarding the underlying common semantic knowledge, such as label hierarchy, label visual similarity, and linguistic/context correlations. More recently, some techniques are explored to capture the human structure information by resorting to complex graphical models (\eg, Conditional Random Fields (CRFs))~\cite{chen2016deeplab}, self-supervised loss~\cite{Gong_2017_CVPR} or human pose priors~\cite{dong2014towards,Fang_2018_CVPR,liang2018look}. However, they did not explicitly model the semantic correlations of different body parts and clothing accessories, and still show unsatisfactory results for rare fine-grained labels. 

One key factor of designing a universal human parsing agent is to have proper transfer learning and knowledge integration among different human parsing tasks, as the label discrepancy across different datasets~\cite{chen2014detect,Gong_2018_ECCV,Gong_2017_CVPR,zhao2018understanding} largely hinders direct data and model unification. In this paper, we achieve this goal by explicitly incorporating human knowledge and label taxonomy into intermediate graph representation learning beyond local convolutions, called ``Graphonomy'' (graph taxonomy). Our Graphonomy learns the global and common semantic coherency in multiple domains via graph transfer learning to solve multiple levels of human parsing tasks and enforce their mutual benefits upon each other. 

\begin{figure}[t]
\begin{center}
   \includegraphics[width=0.9\linewidth]{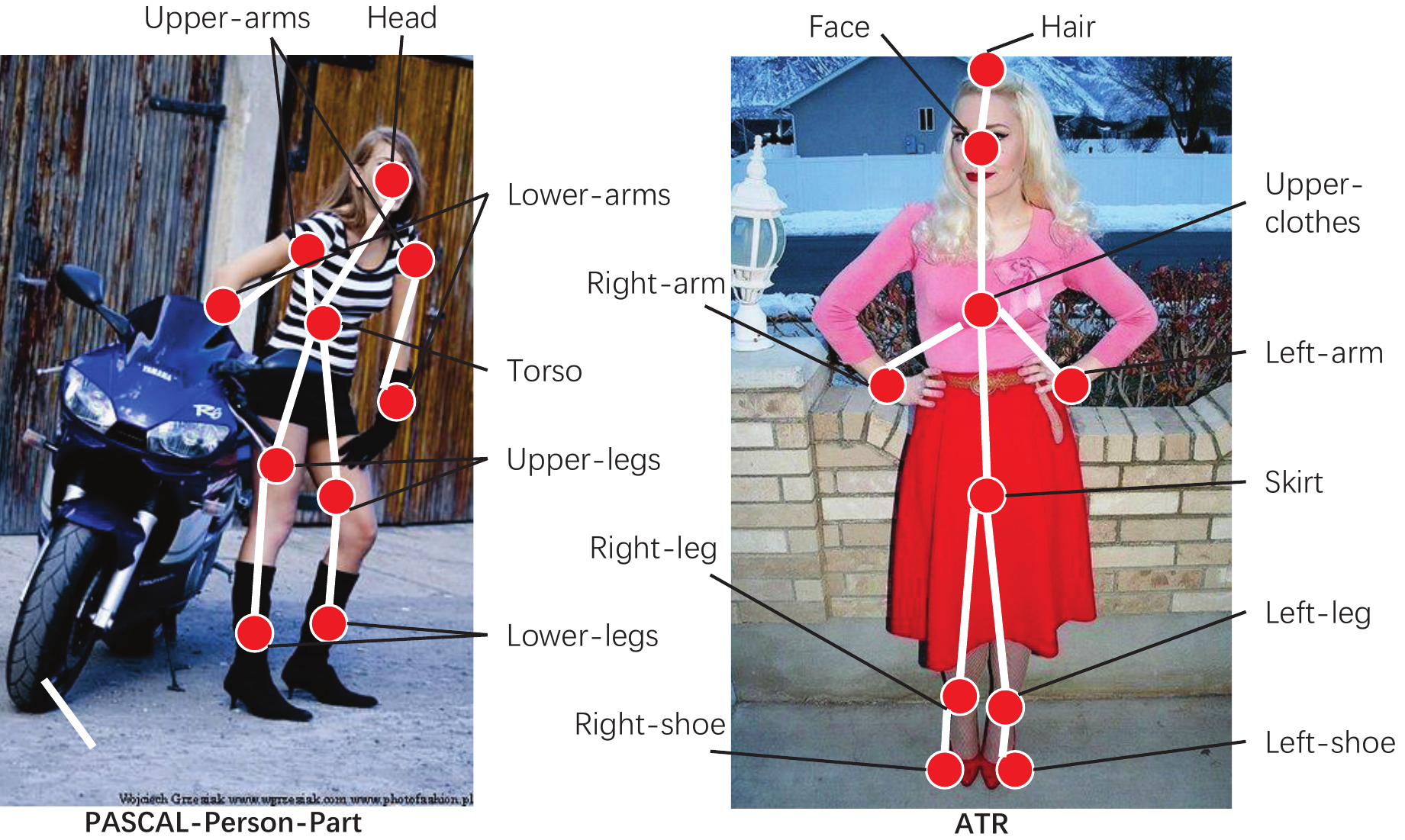}
\end{center}
\vspace{-4mm}
\caption{Examples of the definite connections between each two human body parts, which is the foundation to encode the relations between two semantic nodes in the graph for reasoning. Two nodes are defined related if they are connected by a white line.}
\label{fig:relation}
\vspace{-6mm}
\end{figure}

Taking advantage of geometric deep learning~\cite{kipf2016semi,Lee_2018_CVPR}, our Graphonomy simply integrates two cooperative modules for graph transfer learning. First, we introduce Intra-Graph Reasoning to progressively refine graph representations within the same graph structure, in which each graph node is responsible for segmenting out regions of one semantic part in a dataset. Specifically, we first project the extracted image features into a graph, where pixels with similar features are assigned to the same semantic vertex. We elaborately design the adjacency matrix to encode the semantic relations, constrained by the connection of human body structure, as shown in Fig.~\ref{fig:relation}. After the message propagation via graph convolutions, the updated vertexes are re-projected to make the visual feature maps more discriminative for pixel-level classification.

Additionally, we build an Inter-Graph Transfer module to attentively distill related semantics from the graph in one domain/task to the one in another domain, which bridges the semantic labels from different datasets, and effectively utilize the annotations at multiple levels. To enhance graph transfer capability, we make the first effort to exploit various graph transfer dependencies among different datasets. We encode the relationships between two semantic vertexes from different graphs by computing their feature similarity as well as the semantic similarity encapsulated with linguistic knowledge.

We conduct experiments on three human parsing benchmarks that contain diverse semantic body parts and clothes. The experimental results show that by seamlessly propagating information via Intra-Graph Reasoning and Inter-Graph Transfer, our Graphonomy is able to associate and distill high-level semantic graph representation constructed from different datasets, which effectively improves multiple levels of human parsing tasks.

Our contributions are summarized in the following aspects. 1) We make the first attempts to tackle all levels of human parsing tasks using a single universal model.  In particular, we introduce Graphonomy, a new Universal Human Parsing agent that incorporates hierarchical graph transfer learning upon the conventional parsing network to predict all labels in one system without piling up the complexity. 2) We explore various graph transfer dependencies to enrich graph transfer capability, which enables our Graphonomy to distill universal semantic graph representation and enhance individualized representation for each label graph. 3) We demonstrate the effectiveness of Graphonomy on universal human parsing, showing that it achieves the state-of-the-art results on three human parsing datasets.

\section{Related Work}
\textbf{Human Parsing.}
Human parsing has recently attracted a huge amount of interests and achieved great progress with the advance of deep convolutional neural networks and large-scale datasets. Most of the prior works focus on developing new structures and auxiliary information guidance to improve general feature representation, such as dilated convolution~\cite{chen2016deeplab,yu2015multi}, LSTM structure~\cite{Liang_2017_CVPR,liang2016semantic,liang2015semantic}, encoder-decoder architecture~\cite{chen2018encoder}, and human pose constraints~\cite{Fang_2018_CVPR,liang2018look,Xia_2017_CVPR}. Although these methods show promising results on each human parsing dataset, they directly use one flat prediction layer to classify all labels, which disregards the intrinsic semantic correlations across concepts and utilize the annotations in an inefficient way. Moreover, the trained model cannot be directly applied to another related task without heavy fine-tuning. In this paper, we investigate universal human parsing via graph transfer learning, where each graph encodes a set of concepts in the taxonomy, and all graphs constructed from different datasets are connected following the transfer dependencies to enforce semantic feature propagation.

\textbf{Multi-task Learning.}
Aiming at developing systems that can provide multiple outputs simultaneously for an input, multi-task learning has experienced great progress ~\cite{Dai_2016_CVPR,Dvornik_2017_ICCV,Gong_2018_ECCV,liang2018look,Xia_2017_CVPR,xiao2018unified}. For example,  Gong~\etal~\cite{Gong_2018_ECCV} jointly optimized semantic part segmentation and instance-aware edge detection in an end-to-end way and makes these two correlated tasks mutually beneficial. Xiao~\etal~\cite{xiao2018unified} introduced a multi-task network and training strategy to handle heterogeneous annotations for unified perceptual scene parsing. However, these approaches simply create several branches for different tasks respectively, without exploring explicit relationships among the correlated tasks. In contrast to the existing multi-task learning pipelines, we explicitly model the relations among different label sets and extract a unified structure for universal human parsing via graph transfer learning.

\textbf{Knowledge-guided Graph Reasoning.} Many research efforts recently model domain knowledge as a graph for mining correlations among labels or objects in images, which has been proved effective in many tasks~\cite{Chen_2018_CVPR,kipf2016semi,Lee_2018_CVPR,Liang_2018_CVPR,Wang_2018_CVPR}. For example, Chen~\etal~\cite{Chen_2018_CVPR} leveraged local region-based reasoning and global reasoning to facilitate object detection. Liang~\etal~\cite{Liang_2018_CVPR} explicitly constructed a semantic neuron graph network by incorporating the semantic concept hierarchy. On the other hand, there are some sequential reasoning models for relationships~\cite{Chen_2017_ICCV,li2017attentive}. In these works, a fixed graph is usually considered, while our Graphonomy makes further efforts from external knowledge embedding to graph representation transfer.

\textbf{Transfer Learning.} 
Our approach is also related to transfer learning~\cite{pan2010survey}, which bridges different domains or tasks to mitigate the burden of manual labeling. LSDA~\cite{hoffman2014lsda} transformed whole-image classification parameters into object detection parameters through a domain adaptation procedure. Hu~\etal~\cite{Hu_2018_CVPR} considered transferring knowledge learned from bounding box detection to instance segmentation. Our method transfers high-level graph representations in order to reduce the label discrepancy across different datasets.

\begin{figure*}[t]
\centering
  \includegraphics[width=1.0\linewidth]{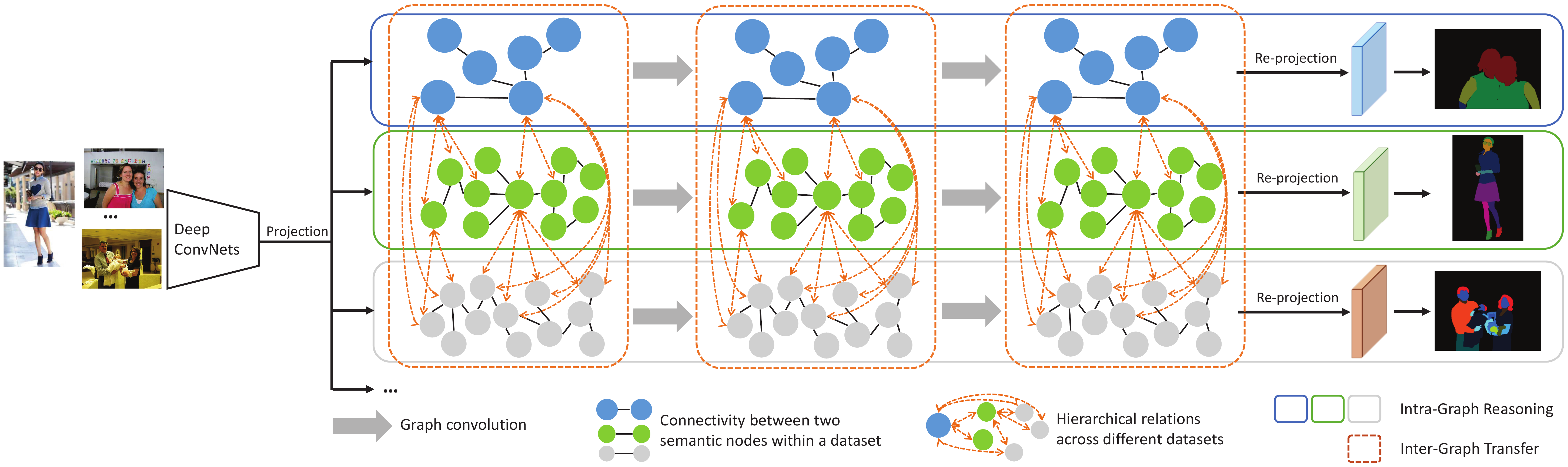}
\vspace{-4mm}
\caption{Illustration of our Graphonomy that tackles universal human parsing via graph transfer learning to achieve multiple levels of human parsing tasks and better annotation utilization. The image features extracted by deep convolutional networks are projected into a high-level graph representation with semantic nodes and edges defined according to the body structure. The global information is propagated via Intra-Graph Reasoning and re-projected to enhance the discriminability of visual features. Further, we transfer and fuse the semantic graph representations via Inter-Graph Transfer driven by hierarchical label correlation to alleviate the label discrepancy across different datasets. During training, our Graphonomy takes advantage of annotated data with different granularity. For inference, our universal human parsing agent generates different levels of human parsing results taking an arbitrary image as input.}
\label{fig:framework}
\vspace{-4mm}
\end{figure*}

\section{Graphonomy}

In order to unify all kinds of label annotations from different resources and tackle different levels of human parsing needs in one system, we aim at explicitly incorporating hierarchical graph transfer learning upon the conventional parsing network to compose a universal human parsing model, named as Graphonomy. Fig.~\ref{fig:framework} gives an overview of our proposed framework. Our approach can be embedded in any modern human parsing system by enhancing its original image features via graph transfer learning. We first learn and propagate compact high-level semantic graph representation within one dataset via Intra-Graph Reasoning, and then transfer and fuse the semantic information across multiple datasets via Inter-Graph Transfer driven by explicit hierarchical semantic label structures.

\subsection{Intra-Graph Reasoning}

Given local feature tensors from convolution layers, we introduce Intra-Graph Reasoning to enhance local features, by leveraging global graph reasoning with external structured knowledge. To construct the graph, we first summarize the extracted image features into high-level representations of graph nodes. The visual features that are correlated to a specific semantic part (\eg, \textit{face}) are aggregated to depict the characteristic of its corresponding graph node. 

Firstly, We define an undirected graph as $G = (V,E)$ where $V$ denotes the vertices, $E$ denotes the edges, and $N = |V|$.
Formally, we use the feature maps $X \in \mathbb{R}^{H \times W \times C}$ as the module inputs, where $H$, $W$ and $C$ are height, width and channel number of the feature maps. We first produce high-level graph representation $Z \in \mathbb{R}^{N \times D}$ of all $N$ vertices, where $D$ is the desired feature dimension for each $v \in V$, and the number of nodes $N$ typically corresponds to the number of target part labels of a dataset.
Thus, the projection can be formulated as the function $\phi$:
\begin{equation}
Z = \phi(X, W),
\label{projection}
\end{equation} 
where $W$ is the trainable transformation matrix for converting each image feature $x_i \in X$ into the dimension $D$.

Based on the high-level graph feature $Z$, we leverage semantic constraints from the human body structured knowledge to evolve global representations by graph reasoning. We introduce the connections between the human body parts to encode the relationship between two nodes, as shown in Fig~\ref{fig:relation}. For example, \textit{hair} usually appears with the \textit{face} so these two nodes are linked. While the \textit{hat} node and the \textit{leg} node are disconnected because they have nothing related. 

Following Graph Convolution~\cite{kipf2016semi}, we perform graph propagation over representations $Z$ of all part nodes with matrix multiplication, resulting in the evolved features $Z^e$:
\begin{equation}
Z^e = \sigma(A^eZW^e),
\label{reasoning1}
\end{equation}
where $W^e \in \mathbb{R}^{D \times D}$ is a trainable weight matrix and $\sigma$ is a nonlinear function. The node adjacency weight $a_{v \to v\prime} \in A^e$ is defined according the edge connections in $(v,v\prime) \in E$, which is a normalized symmetric adjacency matrix. To sufficiently propagate the global information, we employ such graph convolution multiple times (3 times in practice).

Finally, the evolved global context can be used to further boost the capability of image representation. Similar to the projection operation (Eq.~\ref{projection}), we again use another transformation matrix to re-project the graph nodes to images features. We apply residual connection~\cite{he2015deep} to further enhance visual representation with the original feature maps $X$. As a result, The image features are updated by the weighted mappings from each graph node that represents different characteristics of semantic parts.

\subsection{Inter-Graph Transfer}
\label{sec:igt}
To attentively distill relevant semantics from one source graph to another target graph, we introduce Inter-Graph Transfer to bridge all semantic labels from different datasets. Although different levels of human parsing tasks have diverse distinct part labels, there are explicit hierarchical correlations among them to be exploited. For example, \textit{torso} label in a dataset includes \textit{upper-clothes} and \textit{pants} in another dataset, and the \textit{upper-clothes} label can be composed of more fine-grained categories (\eg, \textit{coat}, \textit{T-shirt} and \textit{sweater}) in the third dataset, as shown in Fig.~\ref{fig:graphonomy}. We make efforts to explore various graph transfer dependencies between different label sets, including feature-level similarity, handcraft relationship, and learnable weight matrix. Moreover, considering that the complex relationships between different semantic labels are arduous to capture from limited training data, we employ semantic similarity that is encapsulated with linguistic knowledge from word embedding~\cite{pennington2014glove} to preserve the semantic consistency in a scene. We encode these different types of relationships into the network to enhance the graph transfer capability. 

Let $G_s = (V_s,E_s)$ denotes a source graph and $G_t = (V_t,E_t)$ denotes a target graph, where $G_s$ and $G_t$ may have different structures and characteristics.
We can represent a graph as a matrix $Z \in \mathbb{R}^{N \times D}$, where $N = |V|$ and D is the dimension of each vertex $v \in V$.
The graph transformer can be formulated as:
\begin{equation}
Z_t = Z_t + \sigma(A_{tr}Z_{s}W_{tr}),
\label{reasoning2}
\end{equation}
where $A_{tr}\in \mathbb{R}^{N_t\times N_s}$ is a transfer matrix for mapping the graph representation from $Z_s$ to $Z_t$. $W_{tr}\in \mathbb{R}^{D_s\times D_t}$ is a trainable weight matrix. We seek to find a better graph transfer dependency $A_{tr}=a_{i,j,~i=[1,N_t],~j=[1,N_s]}$, where $a_{i,j}$ means the transfer weight from the $j^{th}$ semantic node of source graph to the $i^{th}$ semantic node of target graph. We consider and compare four schemes for the transfer matrix. 

\textbf{Handcraft relation.} Considering the inherent correlation between two semantic parts, we first define the relation matrix as a hard weight, \ie, $\{0,1\}$. When two nodes have a subordinate relationship, the value of edge between them is 1, else is 0. For example, \textit{hair} is a part of \textit{head}, so the edge value between \textit{hair} node of the target graph and the \textit{head} node of the source graph is 1.

\textbf{Learnable matrix.} In this way, we randomly initialize the transfer matrix $A_{tr}$, which can be learned with the whole network during training.

\textbf{Feature similarity.} The transfer matrix can also be dynamically established by computing the similarity between the source graph nodes and target graph nodes, which have encoded high-level semantic information. The transfer weight $a_{i,j}$ can be calculated as:
\begin{equation}
    a_{i,j} = \frac{exp(sim(v^{s}_i,v^{t}_j))}{\sum_{j} exp(sim(v^{s}_i,v^{t}_j))},
\end{equation}
where $sim(x,y)$ is the cosine similarity between $x$ and $y$. $v^{s}_i$ is the features of the $i^{th}$ target node, and $v^{t}_j$ is the features of the $j^{th}$ source node.

\textbf{Semantic similarity.} Besides the visual information, we further explore the linguistic knowledge to construct the transfer matrix. We use the word2vec model~\cite{pennington2014glove} to map the semantic word of labels to a word embedding vector. Then we compute the similarity between the nodes of the source graph $V_s$ and the nodes of the target graph $V_t$, which can be formulated as:
\begin{equation}
a_{i,j} = \frac{exp(s_{ij})}{\sum_{j} exp(s_{ij})},
\end{equation}
where $s_{ij}$ means the cosine similarity between the word embedding vectors of  $i^{th}$ target node and $j^{th}$ source node.


With the well-defined transfer matrix, the target graph features and source graph knowledge can be combined and propagated again by graph reasoning, the same as the Eq.~\ref{reasoning2}. Furthermore, the direction of the transfer is flexible, that is, two graphs can be jointly transferred from each other. Accordingly, the hierarchical information of different label sets can be associated and propagated via the cooperation of Intra-Graph Reasoning and Inter-Graph Transfer, which enables the whole network to generate more discriminative features to perform fine-grained pixel-wise classification.

\subsection{Universal Human Parsing}
As shown in Fig.~\ref{fig:framework}, apart from improving the performance of one model by utilizing the information transferred from other graphs, our Graphonomy can also be naturally used to train a universal human parsing task for combining diverse parsing datasets. As different datasets have large label discrepancy, previous parsing works must tune highly-specific models for each dataset or perform multi-task learning with several independent branches where each of them handles one level of the tasks. By contrast, with the proposed Intra-Graph Reasoning and Inter-Graph Transfer, our Graphonomy is able to alleviate the label discrepancy issues and stabilize the parameter optimization during joint training in an end-to-end way. 

Another merit of our Graphonomy is the ability to extend the model capacity in an online way. Benefiting from the usage of graph transfer learning and joint training strategy, we can dynamically add and prune semantic labels for different purposes (\eg, adding more dataset) while keeping the network structure and previously learned parameters.

\section{Experiments}
In this section, we first introduce implementation details and related datasets. Then, we report quantitative comparisons with several state-of-the-art methods. Furthermore, we conduct ablation studies to validate the effectiveness of each main component of our Graphonomy and present some qualitative results for the perceptual comparison.

\subsection{Experimental Settings}

\textbf{Implementation Details}
We use the basic structure and network settings provided by DeepLab v3+~\cite{chen2018encoder}. Following \cite{chen2018encoder}, we employ the Xception~\cite{Chollet_2017_CVPR_Xception} pre-trained on COCO~\cite{COCO} as our network backbone and $output~stride=16$. The number of nodes in the graph is set according to the number of categories of the datasets, \ie, $N=7$ for Pascal-Person-Part dataset, $N=18$ for ATR dataset, $N=20$ for CIHP dataset. The feature dimension $D$ of each semantic node is 128. The Intra-Graph Reasoning module has three graph convolution layers with ReLU activate function. For Inter-Graph Transfer, we use the pre-trained model on source dataset and randomly initialize the weight of the target graph. Then we perform end-to-end joint training for the whole network on the target dataset.

During training, the 512x512 inputs are randomly resized between 0.5 and 2, cropped and flipped from the images. The initial learning rate is 0.007. Following~\cite{chen2018encoder}, we employ a ``ploy'' learning rate policy. We adopt SGD optimizer with $momentum=0.9$ and weight decay of $5e-4$. To stabilize the predictions, we perform inference by averaging results of left-right flipped images and multi-scale inputs with the scale from 0.50 to 1.75 in increments of 0.25.

Our method is implemented by extending the Pytorch framework~\cite{paszke2017automatic} and we reproduce DeepLab v3+~\cite{chen2018encoder} following all the settings in its paper. All networks are trained on four TITAN XP GPUs. Due to the GPU memory limitation, the batch size is set to be 12. For each dataset, we train all models at the same settings for 100 epochs for the good convergence. To stabilize the inference, the resolution of every input is consistent with the original image. The code and models are available at \url{https://github.com/Gaoyiminggithub/Graphonomy}.

\begin{table}[]
\centering
\begin{tabular}{cc}
\toprule[0.7pt]
   Method                                                    & Mean IoU(\%)  \\ \hline 
   LIP~\cite{Gong_2017_CVPR}                                 &  59.36 \\
   Structure-evolving LSTM~\cite{Liang_2017_CVPR}            &  63.57 \\
   DeepLab v2~\cite{chen2016deeplab}                         &  64.94 \\
   Li~\etal~\cite{li2017holistic}                            &  66.3  \\
   Fang~\etal~\cite{Fang_2018_CVPR}                          &  67.60  \\ 
   PGN~\cite{Gong_2018_ECCV}                                 &  68.4  \\
   RefineNet~\cite{Lin_2017_CVPR}                            &  68.6  \\
   Bilinski~\etal~\cite{Bilinski_2018_CVPR}                  &  68.6  \\   \hline
   DeepLab v3+~\cite{chen2018encoder}                        &  67.84  \\ 
   Multi-task Learning                                       &  68.13 \\
   Graphonomy (CIHP)                        &  \textbf{71.14} \\
   Graphonomy (Universal Human Parsing)                      &   69.12     \\
\toprule[0.7pt]
\end{tabular}
\caption{Comparison of human parsing performance with several state-of-the-art methods on PASCAL-Person-Part dataset~\cite{chen2014detect}.}
\vspace{-2mm}
\label{tab: pascal}
\end{table}

\begin{table}[]
\centering
\scriptsize
\begin{tabular}{ccc}
\toprule[0.7pt]
   Method                                         & Overall accuracy (\%)   & F-1 score (\%)  \\ \hline 
   LG-LSTM~\cite{liang2015semantic}               &   97.66  &   86.94 \\
   Graph LSTM~\cite{liang2016semantic}            &   98.14  &   89.75  \\
   Structure-evolving LSTM~\cite{Liang_2017_CVPR} &   98.30  &   90.85  \\  \hline 
   DeepLab v3+~\cite{chen2018encoder}             &   97.30  &   84.50   \\
   Multi-task Learning                            &   97.40   &   90.16   \\
   Graphonomy (PASCAL)                            &   \textbf{98.32}  &   \textbf{90.89}  \\
   Graphonomy (Universal Human Parsing)           &   97.69  &   90.16  \\
\toprule[0.7pt]
\end{tabular}
\caption{Human parsing results on ATR dataset~\cite{Co-CNN}.}
\vspace{-2mm}
\label{tab: atr}
\end{table}

\begin{table}[]
\centering
\scriptsize
\begin{tabular}{ccc}
\toprule[0.7pt]
   Method                                         & Mean accuracy(\%)   & Mean IoU(\%)  \\ \hline 
   PGN~\cite{Gong_2018_ECCV}                      &   64.22         &  55.80  \\      \hline
   DeepLab v3+~\cite{chen2018encoder}            &     65.06           &   57.13    \\ 
   Multi-task Learning                           &     65.27           &   57.35   \\
   Graphonomy (PASCAL)                           &     \textbf{66.65}      &  \textbf{58.58} \\
   Graphonomy (Universal Human Parsing)           &     65.73    &   57.78     \\
\toprule[0.7pt]
\end{tabular}
\caption{Performance comparison with state-of-the-art methods on CIHP dataset~\cite{Gong_2018_ECCV}.}
\vspace{-4mm}
\label{tab: cihp}
\end{table}

\begin{table*}[]
\centering
\scriptsize
\tabcolsep 0.011in 
\begin{tabular}{c|c|c|c|c|c|c|c|c|c}
\toprule[0.9pt]
\multirow{2}{*}{~~~\#~~~} & 
\multirow{2}{*}{Basic network~\cite{chen2018encoder}} & \multirow{2}{*}{Adjacency matrix $A^e$} & \multirow{2}{*}{Intra-Graph Reasoning} & \multirow{2}{*}{Pre-trained on CIHP} & \multicolumn{4}{c|}{Inter-Graph Transfer} & \multirow{2}{*}{Mean IoU(\%)} \\ \cline{6-9}
               &                     &                       &                       &                      & Handcraft relation  & Learnable matrix & Feature similarity & Semantic similarity &                   \\  \hline 
            1  &   \checkmark        &       -               &       -               &       -              &       -              &       -          &        -           &        -            &       67.84       \\  \hline 
            2  &   \checkmark        &       -               &   \checkmark          &       -              &       -              &       -          &        -           &        -            &       67.89       \\  \hline 
            3  &   \checkmark        &      \checkmark       &   \checkmark          &       -              &       -              &       -          &        -           &        -            &       68.34       \\  \hline
            4  &   \checkmark        &       -               &       -               &    \checkmark        &       -              &       -          &        -           &        -            &       70.33       \\  \hline
            5  &   \checkmark        &      \checkmark       &   \checkmark          &     \checkmark       &       -              &       -          &        -           &        -            &       70.47       \\  \hline 
            6  &   \checkmark        &      \checkmark       &   \checkmark          &   \checkmark         &    \checkmark        &       -          &        -           &        -            &       70.22       \\  \hline 
            7  &   \checkmark        &      \checkmark       &   \checkmark          &   \checkmark         &       -              &    \checkmark    &        -           &        -            &       70.94       \\  \hline
            8  &   \checkmark        &      \checkmark       &   \checkmark          &   \checkmark         &       -              &       -          &     \checkmark     &        -            &       71.05       \\  \hline
            9  &   \checkmark        &      \checkmark       &   \checkmark          &    \checkmark        &       -              &       -          &        -           &    \checkmark       &       70.95       \\  \hline
            10 &   \checkmark        &      \checkmark       &   \checkmark          &    \checkmark        &       -              &       -          &   \checkmark       &    \checkmark       &  \textbf{71.14}       \\  \hline
            11 &   \checkmark        &     \checkmark        &   \checkmark          &     \checkmark       &       -              &     \checkmark   &     \checkmark     &     \checkmark      &       70.87       \\  \hline
            12 &   \checkmark        &      \checkmark       &    \checkmark         &    \checkmark        &    \checkmark        &    \checkmark    &    \checkmark      &    \checkmark       &       70.69       \\  

\toprule[0.9pt]
\end{tabular}
\vspace{-2mm}
\caption{Ablation experiments on on PASCAL-Person-Part dataset~\cite{chen2014detect}.}
\vspace{-6mm}
\label{tab:ablation}
\end{table*}

\begin{table}[]
\centering
\begin{tabular}{c|c|c}
\toprule[0.7pt]
training data   & Fine-tune  & Graphonomy   \\ \hline
50\%                             &          68.45     &      70.03       \\
80\%                             &          70.02     &      70.26       \\
100\%                            &          70.33     & \textbf{71.14}       \\ 
\toprule[0.7pt]
\end{tabular}
\caption{Evaluation results of our Graphonomy when training on different number of data on PASCAL-Person-Part dataset~\cite{chen2014detect}, in terms of Mean IoU(\%).}
\vspace{-4mm}
\label{tab: few}
\end{table}

\textbf{Dataset and Evaluation Metric}
We evaluate the performance of our Graphonomy on three human parsing datasets with different label definition and annotations, including PASCAL-Person-Part dataset~\cite{chen2014detect}, ATR dataset~\cite{Co-CNN}, and Crowd Instance-Level Human Parsing (CIHP) dataset~\cite{Gong_2018_ECCV}. The part labels among them are hierarchically correlated and the label granularity is from coarse to fine. Referring to their dataset papers, we use the evaluation metrics including accuracy, the standard intersection over union (IoU) criterion, and average F-1 score.

\subsection{Comparison with state-of-the-arts}
\textbf{PASCAL-Person-Part} dataset~\cite{chen2014detect} is a set of additional annotations for PASCAL-VOC-2010~\cite{everingham2010pascal}. It goes beyond the original PASCAL object detection task by providing pixel-wise labels for six human body parts, \ie, \textit{head}, \textit{torso}, \textit{upper-arms}, \textit{lower-arms}, \textit{upper-legs}, \textit{lower-legs}. There are 3,535 annotated images in the dataset, which is split into separate training set containing 1,717 images and test set containing 1,818 images.

We report the human parsing results compared with the state-of-the-art methods in Table~\ref{tab: pascal}. ``Graphonomy (CIHP)'' is the method that transfers the semantic graph constructed on the CIHP dataset to enhance the graph representation on the PASCAL-Person-Part dataset. Some previous methods achieve high performance with over 68\% Mean IoU, thanks to the wiper or deeper architecture~\cite{Bilinski_2018_CVPR,Lin_2017_CVPR}, and multi-task learning~\cite{Gong_2018_ECCV}. Although our basic network (DeepLab v3+~\cite{chen2018encoder}) is not the best, the performance is improved by our graph transfer leaning, which explicitly incorporates human knowledge and label taxonomy into intermediate graph representation, then propagates and updates the global information driven by hierarchical label correlation.

\textbf{ATR} dataset~\cite{Co-CNN} aims to predict every pixel with 18 labels: \textit{face}, \textit{sunglass}, \textit{hat}, \textit{scarf}, \textit{hair}, \textit{upper-clothes}, \textit{left-arm}, \textit{right-arm}, \textit{belt}, \textit{pants}, \textit{left-leg}, \textit{right-leg}, \textit{skirt}, \textit{left-shoe}, \textit{right-shoe}, \textit{bag} and \textit{dress}. Totally, 17,700 images are included in the dataset, with 16,000 for training, 1,000 for testing and 700 for validation.

We report the human parsing results on ATR dataset compared with the state-of-the-art methods in Table~\ref{tab: atr}. ``Graphonomy (PASCAL)'' denotes the method that transfer the high-level graph representation on PASCAL-Person-Part dataset to enrich the semantic information.
Some previous works~\cite{Liang_2017_CVPR,liang2016semantic,liang2015semantic} use the LSTM architecture to improve the performance. Instead, we use the graph structure to propagate and update the high-level information. The advanced results demonstrate that our Graphonomy has stronger capability to learn and enhance the feature representations.

\textbf{CIHP} dataset~\cite{Gong_2018_ECCV} is a new large-scale benchmark for human parsing task, including 38,280 images with pixel-wise annotations on 19 semantic part labels. The images are collected from the real-world scenarios, containing persons appearing with challenging poses and viewpoints, heavy occlusions, and in a wide range of resolutions. Following the benchmark, we use 28,280 images for training, 5,000 images for validation and 5,000 images for testing.

The human parsing results evaluated on CIHP dataset is reported in Table~\ref{tab: cihp}. The previous work ~\cite{Gong_2018_ECCV} achieve high performance with 55\% Mean IoU in this challenging dataset by using multi-task learning. Our Graphonomy (PASCAL) improves the results up to 58.58\%, which demonstrates its superiority and capability to takes full advantages of semantic information to boost the human parsing performance.

\begin{figure}[t]
\begin{center}
   \includegraphics[width=0.9\linewidth]{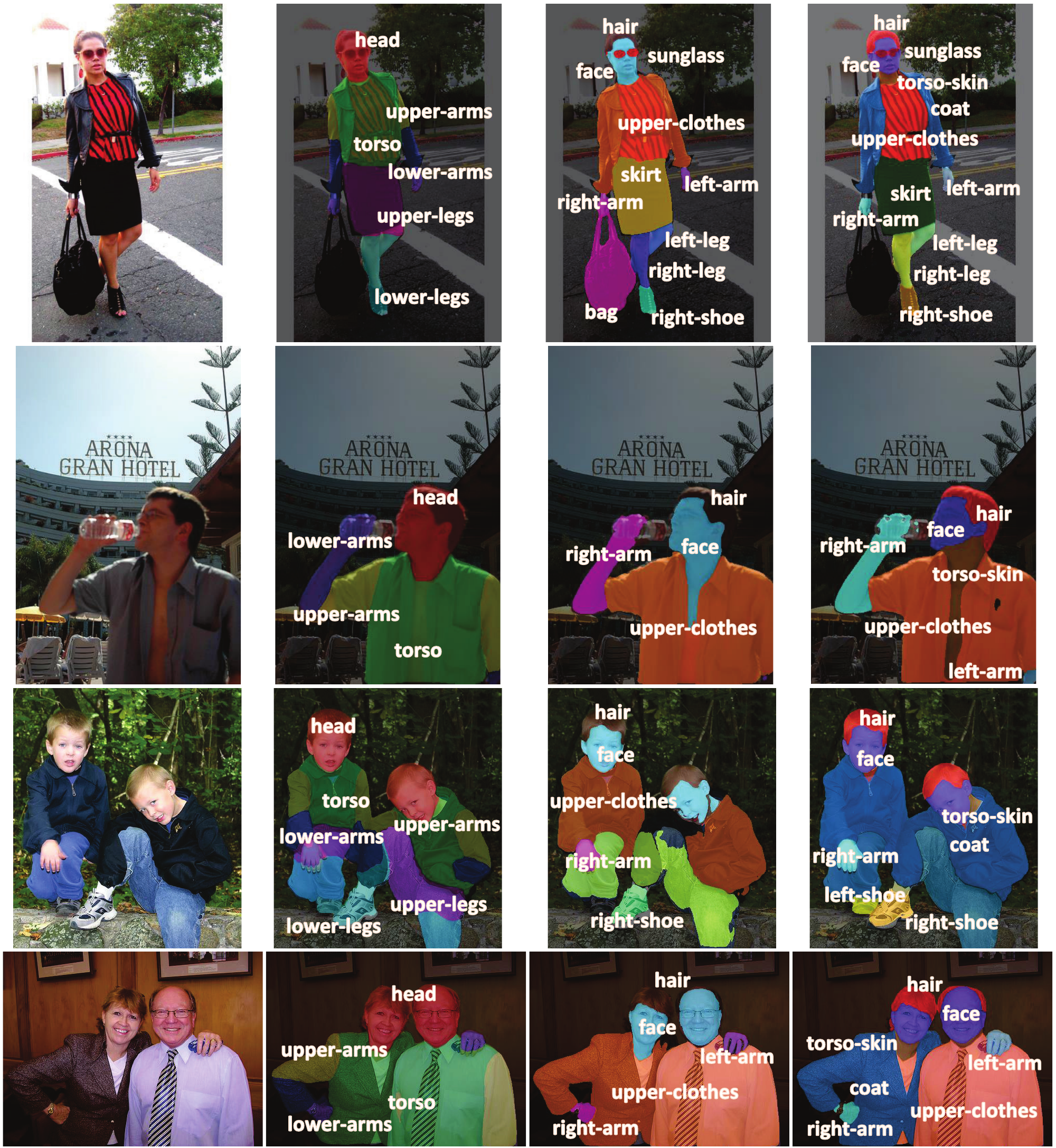}
\end{center}
\vspace{-5mm}
\caption{Examples of different levels of human parsing results generated by our universal human parsing agent, Graphonomy.}
\label{fig:universal}
\vspace{-7mm}
\end{figure}

\subsection{Universal Human Parsing}
To sufficiently utilize all human parsing resources and unify label annotations from different domains or at various levels of granularity, we train a universal human parsing model to unify all kinds of label annotations from different resources and tackle different levels of human parsing, which is denoted as ``Graphonomy (Universal Human Parsing)''. We combine all training samples from three datasets and select images from the same dataset to construct one batch at each step. As reported in Table~\ref{tab: pascal},~\ref{tab: atr},~\ref{tab: cihp}, our method achieves favorable performance on all datasets. We also compare our Graphonomy with multi-task learning method by appending three parallel branches upon the backbone with each branch predicting the labels of one dataset respectively. Superior to multi-task learning, our Graphonomy is able to distill universal semantic graph representation and enhance individualized representation for each label graph. 

We also present the qualitative universal human parsing results in Fig.~\ref{fig:universal}. Our Graphonomy is able to generate precise and fine-grained results for different levels of human parsing tasks by distilling universal semantic graph representation to each specific task, which further verifies the rationality of our Graphonomy based on the assumption that incorporating hierarchical graph transfer learning upon the deep convolutional networks can capture the critical information across the datasets to achieve good capability in universal human parsing.

\begin{figure*}[t]
\centering
\includegraphics[width=0.9\linewidth]{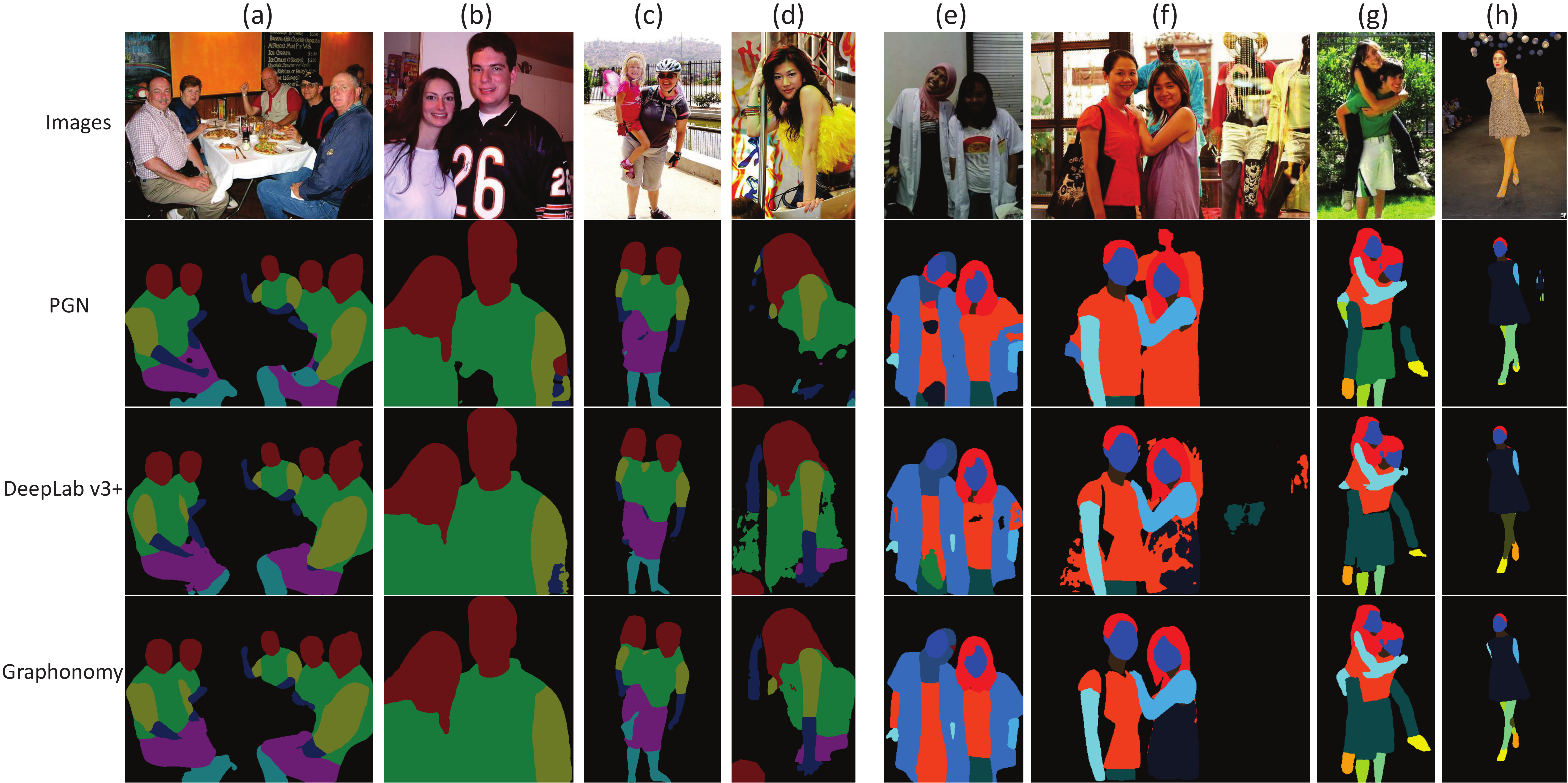}
\vspace{-1mm}
\caption{Visualized comparison of human parsing results on PASCAL-Person-Part dataset~\cite{chen2014detect} (Left) and CIHP dataset~\cite{Gong_2018_ECCV} (Right).}
\label{fig:visual}
\vspace{-6mm}
\end{figure*}

\subsection{Ablation Studies}
We further discuss and validate the effectiveness of the main components of our Graphonomy on PASCAL-Person-Part dataset~\cite{chen2014detect}.

\textbf{Intra-Graph Reasoning.} As reported in Table~\ref{tab:ablation}, by encoding human body structure information to enhance the semantic graph representation and propagation, our Intra-Graph Reasoning acquires 0.50\% improvements compared with the basic network (\#1 vs \#3). To validate the significance of adjacency matrix $A^e$, which is defined according to the connectivity between human body parts and enables the semantic messages propagation, we compare our methods with and without $A^e$ (\#2 vs \#3). The comparison result shows that the human prior knowledge makes a larger contribution than the extra network parameters brought by the graph convolutions.

\textbf{Inter-Graph Transfer.} To utilize the annotated data from other datasets, previous human parsing methods must be pre-trained on the other dataset and fine-tuned on the evaluation dataset, as the \#4 result in Table~\ref{tab:ablation}. Our Graphonomy provides a Inter-Graph Transfer module for better cross-domain information sharing. We further compare the results of difference graph transfer dependencies introduced in Section~\ref{sec:igt}, to find out the best transfer matrix to enhance graph representations. Interestingly, it is observed that transferring according to handcraft relation (\#6) diminishes the performance and the feature similarity (\#8) is the most powerful dependency. It is reasonable that the label discrepancy of multiple levels of human parsing tasks cannot be solved by simply defining the relation manually and the hierarchical relationship encoded by the feature similarity and semantic similarity is more reliable for information transferring. Moreover, we compare the results of different combinations of the transfer methods, which bring in a little more improvement. In our Graphonomy, we combine feature similarity and semantic similarity for the Inter-Graph Transfer, as more combinations cannot contribute to more improvements.

\textbf{Different number of traning data.} Exploiting the intrinsic relations of semantic labels and incorporating hierarchical graph transfer learning upon the conventional human parsing network, our Graphonomy not noly tackle multiple levels of human praing tasks, but also alleviate the need of heavy annotated traning data to achieve the desired performance. We conduct extensive experiments on transferring the model pre-trained on CIHP dataset to PASCAL-Person-Part dataset. We use different annotated data in training set by random sampling for training and evaluate the models on the whole test set. As summarized in Table~\ref{tab: few}, simply fine-tuning the pre-trained model without our proposed Inter-Graph Transfer obtains 70.33\% mean IoU with all training data. However, our complete Graphonomy architecture uses only 50\% of the training data and achieves comparable performance. With 100\% training data, our approach can even outperforms the fine-tuning baseline for 0.81\% in average IoU. This superior performance confirms the effectiveness of our Graphonomy that seamlessly bridges all semantic labels from different datasets and attains the best utilization of data annotations.

\subsection{Qualitative Results}
The qualitative results on the PASCAL-Person-Part dataset~\cite{chen2014detect} and the CIHP dataset~\cite{Gong_2018_ECCV} are visualized in Fig.~\ref{fig:visual}. As can be observed, our approach outputs more semantically meaningful and precise predictions than other two methods despite the existence of large appearance and position variations. Taking (b) and (e) for example, when parsing the clothes, other methods are suffered from strange fashion style and the big logo on the clothes, which leads to incorrect predictions for some small regions. However, thanks to the effective semantic information propagation by graph reasoning and transferring, our Graphonomy successfully segments out the large clothes regions. More superiorly, with the help of the compact high-level graph representation integrated from different sources, our method generates more robust results and gets rid of the disturbance from the occlusion and background, like (c) and (d). Besides, we also present some failure cases (g) and (h), and find that the overlapped parts and the very small persons cannot be predicted precisely, which indicates more knowledge is desired to be incorporated into our graph structure to tackle the challenging cases.

\section{Conclusion}
In this work, we move forward to resolve all levels of human parsing tasks using a universal model to alleviate the label discrepancy and utilize the data annotation. We proposed a new universal human parsing agent, named as Graphonomy, that incorporates hierarchical graph transfer learning upon the conventional parsing network to predict all labels in one system without piling up the complexity. The solid and consistent human parsing improvements of our Graphonomy on all datasets demonstrates the superiority of our proposed method. The advantageous universal human parsing performance further confirms that our Graphonomy is strong enough to unify all kinds of label annotations from different resources and tackle different levels of human parsing needs. In future, we plan to generalize Graphonomy to more general semantic segmentation tasks and investigate how to embed more complex semantic relationships naturally into the network design.

\section{Acknowledgements}
 This work was supported by the Sun Yat-sen University Start-up Foundation Under Grant No. 76160-18841201, in part by the National Key Research and Development Program of China under Grant No. 2018YFC0830103, in part by National High Level Talents Special Support Plan (Ten Thousand Talents Program), and in part by National Natural Science Foundation of China (NSFC) under Grant No. 61622214, and 61836012.

{\small
\bibliographystyle{ieee_fullname}
\bibliography{egbib}
}

\end{document}